\documentclass{article}

\usepackage{microtype}
\usepackage{graphicx}
\usepackage{subcaption}
\usepackage{booktabs} % for professional tables
\usepackage{amssymb}
\usepackage{adjustbox}
\usepackage{balance}
\usepackage{tabularx}

\usepackage{hyperref}

\usepackage{amsmath}
\newcommand{\norm}[1]{\left\lVert#1\right\rVert_2^2}

\usepackage[accepted]{icml2019}

\icmltitlerunning{DeepChrome 2.0}

\begin{document}

\twocolumn[
\icmltitle{DeepChrome 2.0:\\Investigating and Improving Architectures, Visualizations, \& Experiments
}



\icmlsetsymbol{equal}{*}

\begin{icmlauthorlist}
\icmlauthor{Saurav Kadavath}{equal,to}
\icmlauthor{Samuel Paradis}{equal,to}
\icmlauthor{Jacob Yeung}{equal,to}
\end{icmlauthorlist}

\icmlaffiliation{to}{University of California, Berkeley, USA}

% \icmlcorrespondingauthor{Saurav Kadavath}{sauravkadavath@berkeley.edu}
% \icmlcorrespondingauthor{Samuel Paradis}{sauravkadavath@berkeley.edu}
% \icmlcorrespondingauthor{Jacob Yeung}{sauravkadavath@berkeley.edu}

\icmlkeywords{Machine Learning, ICML}

\vskip 0.3in
]



\printAffiliationsAndNotice{\icmlEqualContribution} % otherwise use the standard text.

\begin{abstract}
Histone modifications play a critical role in gene regulation. Consequently, predicting gene expression from histone modification signals is a highly motivated problem in epigenetics. We build upon the work of \citet{singh2016deepchrome}, who trained classifiers that map histone modification signals to gene expression. We present a novel visualization technique for providing insight into combinatorial relationships among histone modifications for gene regulation that uses a generative adversarial network to generate histone modification signals. We also explore and compare various architectural changes, with results suggesting that the 645k-parameter convolutional neural network from DeepChrome has the same predictive power as a 12-parameter linear network. Results from cross-cell prediction experiments, where the model is trained and tested on datasets of varying sizes, cell-types, and correlations, suggest the relationship between histone modification signals and gene expression is independent of cell type. We release our PyTorch re-implementation of DeepChrome on GitHub \footnote{\url{github.com/ssss1029/gene_expression_294}}.\parfillskip=0pt
\end{abstract}

\parfillskip=0pt

\section{Introduction}

In gene regulation, different processes can augment gene expression. Many mechanisms are used to control this augmentation, but a key mechanism in gene regulation is the modification of histones. DNA is often packaged around histone proteins, and \textit{histone modifications} (HMs) control the structure of this packing. Histone proteins often have regions where events such as methylation or acetylation occur, and the presence or absence of these events are HMs. Many studies have experimentally confirmed that a relationship exists between HMs and gene expression. However, an open problem in this area is \textit{predicting gene expression from HM signals.} This problem is highly motivated, as predicting gene expression from HM signals provides the potential to better understand combinatorial effects in gene regulation, which is critical towards the development of epigenetic drugs.

In this paper, we build upon the work of \citet{singh2016deepchrome}, which maps HM signals to gene expression using a deep convolutional neural network. They also propose an optimization-based visualisation technique to extract an intuitive representation of what the network learned. DeepChrome outperformed state-of-the-art models at the time of its release, and acts as an important stepping stone towards understanding combinatorial effects in gene regulation. We build upon this work in three key ways: (1) developing a novel visualization technique, (2) exploring architectural improvements, and (3) conducting cross-cell prediction experiments.

\citet{singh2016deepchrome} desired to visualize the trained network to understand \textit{how it learned} to map HM signals to gene expression. To do so, the authors take a DeepChrome classifier, choose a target class (either +1 or -1), randomly initialize input $X$, and then use backpropagation and gradient descent to iteratively modify $X$ such that it becomes maximally representative of a particular gene expression label. This allowed the authors to confirm that the reasoning of the network aligned with recent biological findings. However, the authors also claim such a technique allows for learning novel insights into combinatorial relationships among HMs for gene regulation. Our results suggest that rather, such a technique extracts the ``guts" from the network, which do not transfer well into contexts outside of the network. In other words, the input $X$ that results from the optimization-based visualization represents an input that is specifically engineered to maximally activate some neural network; there is nothing constraining $X$ to be a realistic HM signal. We present a visualization technique that generates inputs that both maximally activate our classifier and are realistic HM signals. We use a generative adversarial network \citep{goodfellow2014generative}, or GAN, to generate millions of HM signals, and select the signals that maximally activate the classifier.

We explore and compare various architectural changes, such as maxpooling, averagepooling, strided convolutions, and a simple linear network. While we hypothesize that averagepooling or strided convolutions may capture the local intricacies that maxpooling ignores, we find minimal differences in performance between the three. Since changing architectures seemed to make no difference, we hypothesize the models are possibly not leveraging the full power of the network, and as a result, try to train a 12-parameter linear network.

We also explore the relationship between the impact of HM signals on gene expression and cell type. In \citet{singh2016deepchrome}, the authors focus on training separate DeepChrome models for each cell type in their dataset and briefly mention cross-cell experiments as a future avenue of research. We build upon their work by training and testing on various sets of cell types. Our goal is to understand how effectively the knowledge of HM signals transfers across cell types. We use RNA-seq RPKM counts from the Roadmap Epigenomics Consortium (REMC) database as the basis to calculate correlations between each cell type \citep{remc}. We find there is essentially no relationship between train and test cell correlations and model performance, implying the impact of HM signals on gene expression is independent of cell type.

\subsection{Key Contributions}
\begin{enumerate}
    \item A novel visualization technique based on the well motivated goals of \citet{singh2016deepchrome} that provides insight into the network and the potential to learn novel insights into combinatorial relationships among HMs for gene regulation.
    \item Experimental results suggesting the relationship between HM signals and gene expression is independent of cell type.
    \item Experimental results suggesting signals learned by DeepChrome are only as informative as the raw frequency of each HM.
\end{enumerate}

\section{Related Works}
\citet{limetal2009} perform wet lab experiments testing if there is a relationship between HM marks and gene regulation. The authors use clustering to group genes that have similar expressions, and find that the tested inducible genes typically have active methylation marks. \citet{Karlic} then confirm a high correlation between predicted and observed HM marks. They use a linear model to regress HM levels from the number of promoters. Building on this, \citet{Costa2011} use a mixture of linear regression models. Both of these techniques use the mean signal of the entire transcription start site (TSS) flanking regions as input.

\citet{cheng2011} improve on the data input technique by using a bin specific approach which divides the TSS and TTS (transcription termination site) into 160 bins of 100 base pairs each. For each bin, they extract the mean signal of the chromatin features, and train a support vector machine to classify if the gene was active or not. The authors show that bins further away are less informative than bins closer to the TSS - this is an interesting insight which DeepChrome confirms.

Extending this binning technique, \citet{dong2012} use a Random Forest Classifier on HM signals to classify high and low gene expressions. Then, using the classified outputs, they build a linear regression model for each bin to predict gene expression values. Unlike in previous experiments, the authors select a single model with the highest AUROC. The major downside to this approach is the lack of influences of bins besides the ``best bin".

Expanding on the previous works, \citet{ho2015} apply rule learning to explore combinatorial effects of HMs on gene expression. \citet{ernst2015} take advantage of histone marks, DNA accessibility, and DNA methylation to create an ensemble of regression trees to predict and impute epigenetic signals.

% In summary, experimental results confirming a relationship between HMs and gene expression ignited this series of studies. This relationship was then confirmed with models, with increasing complexity.

\subsection{DeepChrome}
The goal of DeepChrome is to use the information about HMs around the TSS of a gene in order to predict whether or not that gene is expressed. The authors of DeepChrome look at 56 different cell types and 5 different HMs: H3K27me3, H3K36me3, H3K4me1, H3K4me3, and H3K9me3. From REMC \cite{remc}, the authors obtain the HM counts, as well as the normalized median RPKM of RNA-seq experiments per cell in order to binarize gene expression. In order to input these reads into the model, similar to previous works, DeepChrome uses a binning strategy. For a particular HM signal, they focus on the 5000 basepairs before and after the TSS, resulting in 100 bins representing 100 basepairs each. Given 5 HM signals, the authors use this technique to construct a 5$\times$100 matrix, which represents a single input for DeepChrome, and conceptually contains information about which HMs occur around the TSS for a given gene. DeepChrome used a simple architecture: a convolutional layer with a kernel size of 5$\times$10, with the 5 representing the 5 HMs, followed by a temporal maxpooling layer, dropout, 2 feed-forward layers, and a 2-dimensional softmax.

\subsection{AttentiveChrome and DeepDiff}
\citet{singh2017attend} followed up with AttentiveChrome, an attention-based recurrent neural network which uses the same inputs and outputs as DeepChrome. An advantage of AttentiveChrome over previous deep learning attempts is the increased interpretability through visualization of the attention weights. They view the attention maps as heatmaps to show what the network is learning. The authors note that their new visualization technique achieves better correlations between active histone promoter marks and active regions than prior techniques such as class-based optimization and saliency maps. The authors further extend AttentiveChrome with DeepDiff \cite{deepdiff}, a multi-task deep neural network. They find that the learned attention weights match the experimental observations by \citet{gregoire2016} for upregulated and downregulated genes in blood cells and leukemia cells.

\section{Experimental Design}

\subsection{Visualization Techniques}

% STICKING HERE FOR ORDERING
%%%%%%%%%%%%%%%%%%%%%%%%%%%%%%%%%%%%%%%%%%%%%%%%
\begin{figure}[t]
\begin{subfigure}{.155\textwidth}
  \centering
    \includegraphics[width=\textwidth]{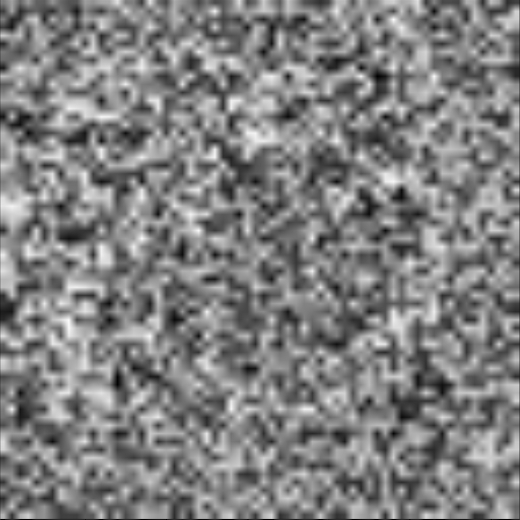}
  \caption{}
  \label{fig:viz-backprop-a}
\end{subfigure}%
\hspace{1pt}
\begin{subfigure}{.155\textwidth}
  \centering
    \includegraphics[width=\textwidth]{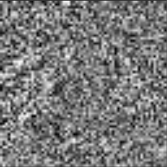}
  \caption{}
  \label{fig:viz-backprop-b}
\end{subfigure}%
\hspace{1pt}
\begin{subfigure}{.155\textwidth}
  \centering
    \includegraphics[width=\textwidth]{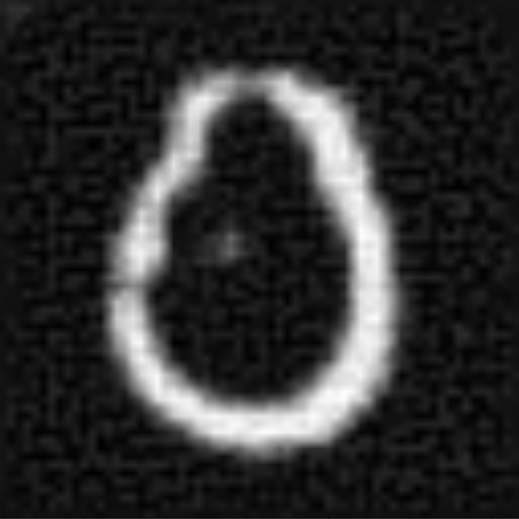}
  \caption{}
  \label{fig:viz-backprop-c}
\end{subfigure}%
\vspace{-6pt}
\caption{\textbf{Backpropagation Visualization Technique.} Visualising results using the backpropagation technique from \citet{singh2016deepchrome} on a classifier trained on MNIST dataset, attempting to find input $X$ that maximizes confidence for label $0$. (a) starts from random $X$ and uses a loss dependent on the classifier, (b) starts from random $X$ and uses a loss dependent on classifier and discriminator, (c) starts from in-distribution $X$ and uses a loss dependent on classifier, discriminator, and deviation from starting $X$.
 }
\label{fig:viz-backprop}
\end{figure} 
\begin{figure}[t]
\begin{subfigure}{.155\textwidth}
  \centering
    \includegraphics[width=\textwidth]{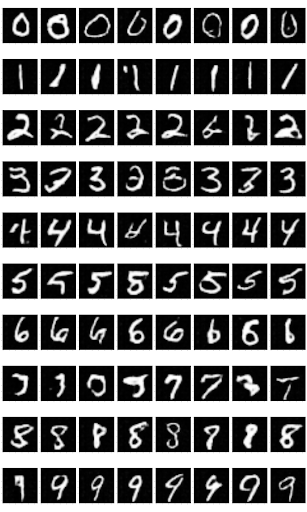}
  \caption{$\scriptstyle{P(f(X) = c) > 0.75}$}
\end{subfigure}%
\hspace{1pt}
\begin{subfigure}{.155\textwidth}
  \centering
    \includegraphics[width=\textwidth]{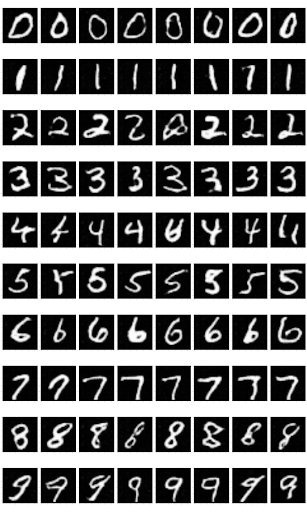}
  \caption{$\scriptstyle{P(f(X) = c) > 0.99}$}
  \end{subfigure}%
\hspace{1pt}
\begin{subfigure}{.155\textwidth}
  \centering
    \includegraphics[width=\textwidth]{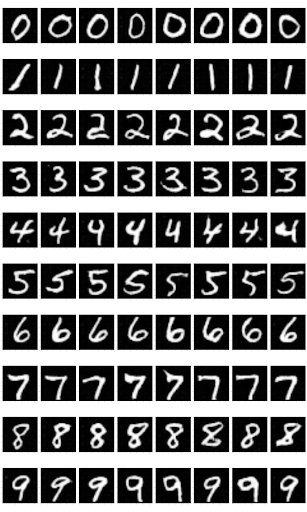}
  \caption{$\scriptstyle{P(f(X) = c) > 0.9999}$}
\end{subfigure}%
\vspace{-6pt}
\caption{\textbf{Monte Carlo GAN Visualization Technique.} Results on the MNIST DeepChrome-style classifier and a GAN; first ten inputs that exceeded required threshold for each class. The model is more confident on images that are in the training distribution, thus the method implicitly selects samples from the GAN that are more realistic. The $0.9999$ threshold shows we are able to generate inputs that both maximally activate the classifier and are within a desired data distribution.
 }
\label{fig:viz-gan-mnist}
\end{figure}
%%%%%%%%%%%%%%%%%%%%%%%%%%%%%%%%%%%%%%%%%%%%%%%%

In \citet{singh2016deepchrome}, the authors use a visualization technique to generate inputs that maximally activate some classifier. To do so, they utilize backpropagation and gradient descent in attempt to find some input $X$ that is most representative of a particular gene expression label. They claim this visualization technique provides the potential to learn novel insights into combinatorial relationships among HMs for gene regulation, but we contest that this class-based optimization, which extracts the ``guts" from the network, does not transfer well into contexts outside of the network. While the backpropagation technique generates inputs that maximally activate some classifier, there are no guarantees the inputs are realistic, which limits any ability to learn novel insights from the network.

In order to combat this, we attempt to generate inputs that both maximally activate some classifier and are within a desired data distribution. We switch to a vision domain to leverage our innate understanding of images to visually confirm the plausibility of the generated inputs. We first exactly matched the DeepChrome architecture in the number and type of parameters, and then trained a classifier on the MNIST dataset \cite{lecunmnisthandwrittendigit2010}. Results from using the backpropagation technique from \citet{singh2016deepchrome} are shown in Fig.~\ref{fig:viz-backprop-a}, where loss $\mathcal{L} = \mathcal{L}_\textnormal{Classifier}(X_t,c)$. In an attempt to constrain the generated input to be within some data distribution, we use a GAN. A trained GAN consists of a generator, which generates inputs to mimic some training data, and a discriminator, which classifies if the generated input is within the same distribution as the training data. This allows us to add a term to our loss pertaining to the output of the discriminator, setting $\mathcal{L} = \mathcal{L}_\textnormal{Classifier}(X_t,c) + \lambda\mathcal{L}_\textnormal{Discriminator}(X_t)$. Thus, when we backpropagate on $X$, we try to both maximize the likelihood of being classified as some class \textit{and} maximize discriminator confidence that its within the training data. As seen in Fig.~\ref{fig:viz-backprop-b}, this technique quickly finds an input that incurs zero loss, but this input still does not look to be from the MNIST dataset. A third idea was to hot-start the input from the generator, and then enforce a loss on deviating from the starting image, idea being it may augment the image in some way representative of high activation inputs. Here, $\mathcal{L} = \mathcal{L}_\textnormal{Classifier}(X_t,c) + \lambda\mathcal{L}_\textnormal{Discriminator}(X_t) + \Phi\norm{X_t - X_1}$. However, for large $\Phi$, the image does not change, and for smaller $\Phi$, the small changes are indistinguishable from random noise, meaning it is not augmenting the image in a meaningful way (see Fig.~\ref{fig:viz-backprop-c}). Results show that during backpropagation, there is nothing enforcing that the generated example maintains any non-adversarial structure, as it quickly finds inputs that are adversarial to each loss term.

In order to generate inputs that both maximally activate some classifier and are within a desired data distribution, another idea is to generate many inputs using the generator from the GAN, and select the inputs that maximally activate some label in the classifier. This represents a Monte Carlo approach. The idea is that, given the GAN generates in-distribution inputs, sampling many inputs from the GAN will provide some inputs that highly activate the classifier, and since we constrained the generated inputs to be reasonable, we know whatever inputs maximally activate the classifier are also reasonable. Results using the MNIST DeepChrome-style classifier and a GAN are shown in Fig.~\ref{fig:viz-gan-mnist}. This technique can directly be applied to the genomics settings, where the classifier maps the  $5\times100$ HM signals to gene expression, and the GAN generates the $5\times100$ HM inputs. Thus, using this probing technique, we can produce inputs that both maximally activate the classifier and represent realistic HM signals.

\subsection{Architectural Modifications}
The original DeepChrome model uses a maxpool to reduce the dimensions of their input. In vision, maxpooling makes intuitive sense since the macro-level information of interest is robust to the information loss incurred by a maxpooling layer. However, in genomics, we hypothesize that pixel-level changes matter because local changes strongly affect gene expression predictions. We hypothesize that averagepooling may capture the local intricacies that maxpooling ignores. In addition, we also experimented with strided convolutions and a simple linear network.

\subsection{Cross-cell Experiment Design}

\begin{figure}[t]
\begin{subfigure}{.155\textwidth}
    \centering
    \includegraphics[width=\textwidth]{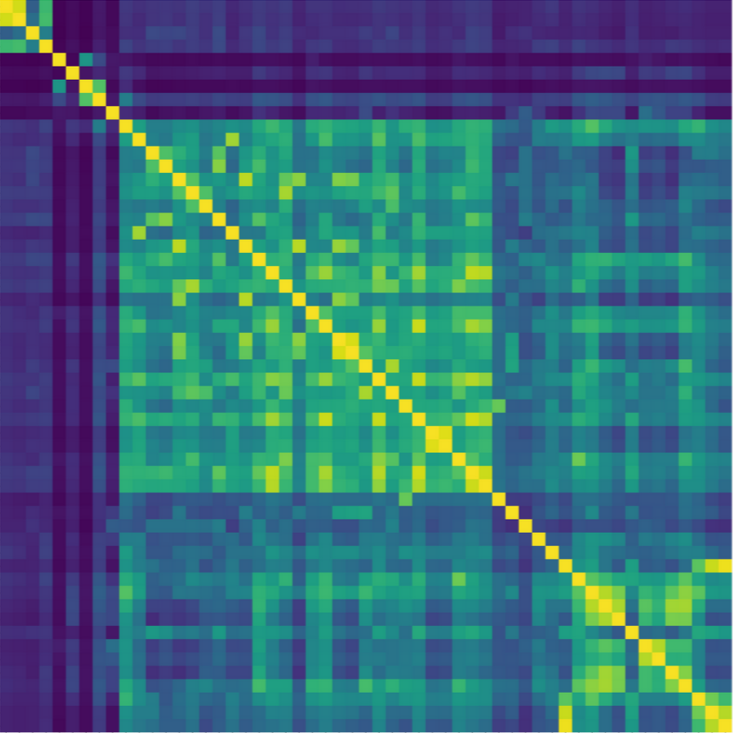}
    \caption{Train}
    \label{fig:train-corr}
\end{subfigure}
\begin{subfigure}{.155\textwidth}
    \centering
    \includegraphics[width=\textwidth]{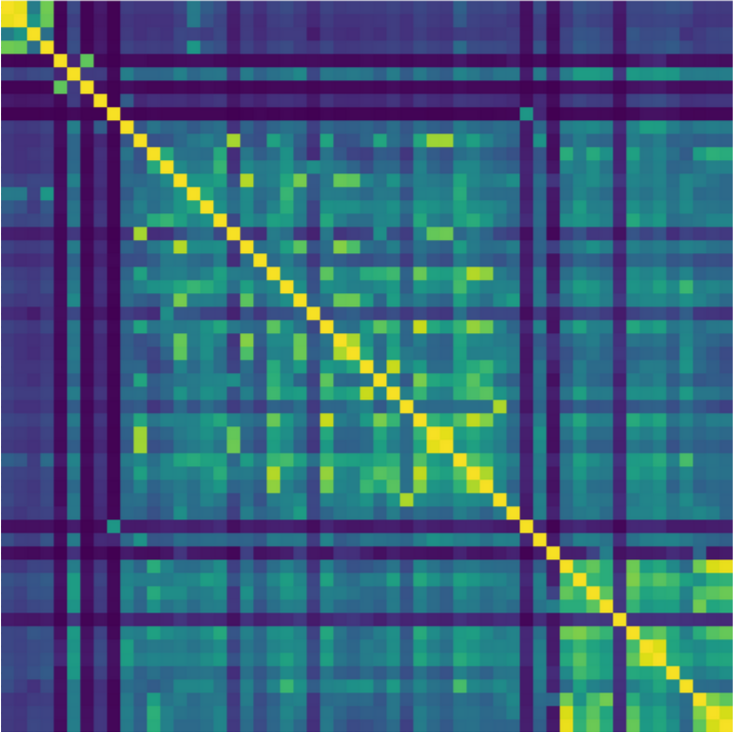}
    \caption{Validation}
    \label{fig:valid-corr}
\end{subfigure}
\begin{subfigure}{.155\textwidth}
    \centering
    \includegraphics[width=\textwidth]{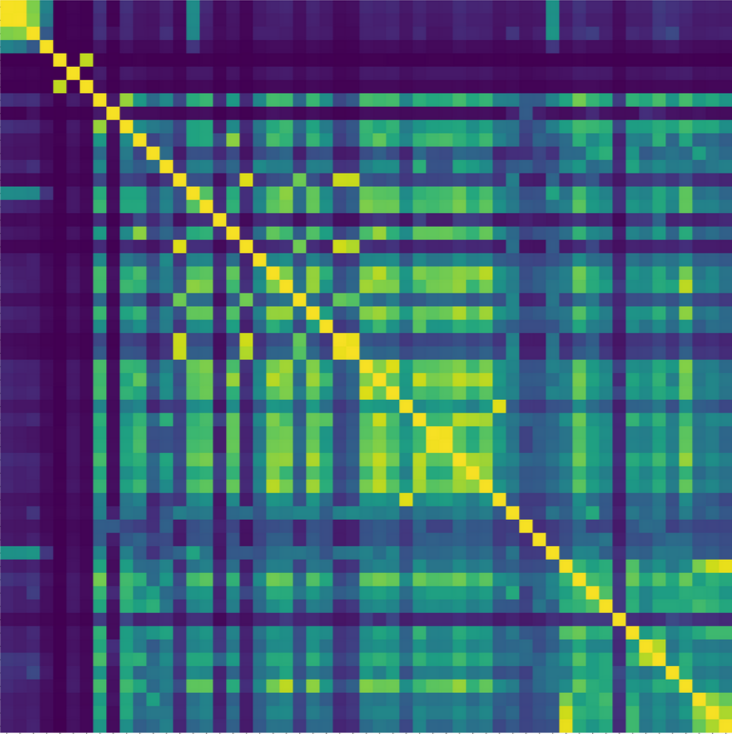}
    \caption{Test}
    \label{fig:test-corr}
\end{subfigure}
% \begin{subfigure}{.0125\textwidth}
%     \centering
%     \includegraphics[width=\textwidth]{figures/corr-bar.png}
%     \caption*{}
%     \label{fig:train-corr}
% \end{subfigure}
\vspace{-6pt}
\caption{\textbf{Gene Expression Correlation Heatmaps.} Correlation heatmaps created from the RNA-seq RPKM counts for all genes in every cell. Brighter colors represent higher correlations. As expected, the highest correlation is along the diagonal, with a correlation of $1$. Overall, the correlations are qualitatively similar across all 3 data splits, despite some incongruencies such as the dark lines in (b) and (c) that are not present in (a).
 }
\label{fig:corr}
\end{figure}

In \citet{singh2016deepchrome}, the authors focus on training separate DeepChrome models for each cell type in their dataset, and they briefly mention cross-cell experiments as a future avenue of research. We build upon their work by exploring training and testing on various sets of cell types. Our goal is to understand how effectively the knowledge of HM signals transfers across cell types.

\textbf{Gene Expression Correlation.}
We hypothesize that cells with similar gene expression profiles would likely benefit the most from cross-cell training. Consequently, we utilize the RNA-seq RPKM counts of each gene for every cell as the basis for determining similarity. We obtain the RNA-seq RPKM counts from the REMC database, with each cell having 19,795 genes \citep{remc}. The genes are then split equally into training, validation, and testing sets based on DeepChrome's datasets. We then calculate correlations between all cell types and normalize the range from 0 to 1. As we see in Fig.~\ref{fig:corr}, there are groups of correlated cells. However, we also observe some incongruencies--we see some cells are not as correlated in the validation and testing sets as they are in the training set (identified by dark lines crossing the validation and testing heatmaps). Despite these inconsistencies, the correlations are qualitatively similar across all 3 dataset splits.

\textbf{Experiment Formulation.}
In order to understand the transferability of HM signals, we conduct and compare several experiments. Given a specific test cell $A$, our baseline experiment (called ``DeepChrome") is to train on data from cell $A$ itself. We also train on all cells (denoted by ``All"), cells with a correlation $>0.75$ with cell $A$ (denoted by ``Highly"), cells with a correlation $>0.5$ with cell $A$ (denoted by ``Somewhat"), and a random subset of 10 cells (denoted by ``Random"). The Random experiments are used as a control group for the other correlation experiments to ensure differences are rooted in correlation, not number of training cells. For each of the All, Highly, Somewhat, and Random categories, we run one experiment where we include the target cell $A$, and one where we exclude it; these are denoted as ``Inclusive" and ``Exclusive" respectively. In order to better understand the transferability of the models, we test each model from our DeepChrome experiments on all 55 other cells that it was not trained on. These are denoted as our Test-on-Rest experiments. For all of our cross-cell experiments, we replace the maxpool with an averagepool.

\subsection{Linear Baseline}
We also include a basic 12-parameter linear baseline model to test the relationship between HMs and gene expression. This network takes in the temporal mean of the original input to DeepChrome, resulting in a 5$\times$1 input, and returns a binary classification of gene expression. A model was trained separately on each cell. We train each linear model 5 times, selecting the model with the highest validation AUROC for testing.

\section{Experimental Results}

\begin{table}[t]
\caption{
\small
\textbf{Genomic GAN Sanity Check}. Checking if DeepChrome data and GAN generated data incur similar outputs from a DeepChrome classifier, with a random input as a control group. Mean over 10k inputs for each.\\}
\centering
\small
\begin{tabular}{| l || c | c |}
\hline
 & Mean Positive Prob. & Mean Negative Prob.
\\ \hline\hline 
DeepChrome & 0.30 & 0.70
\\ \hline
GAN & 0.31 & 0.69
\\ \hline
Random & 0.19 & 0.81
\\ \hline
\end{tabular}
\label{tab:gan-sanity}
\end{table}

\begin{figure}[t]
  \centering
    \includegraphics[width=.47\textwidth]{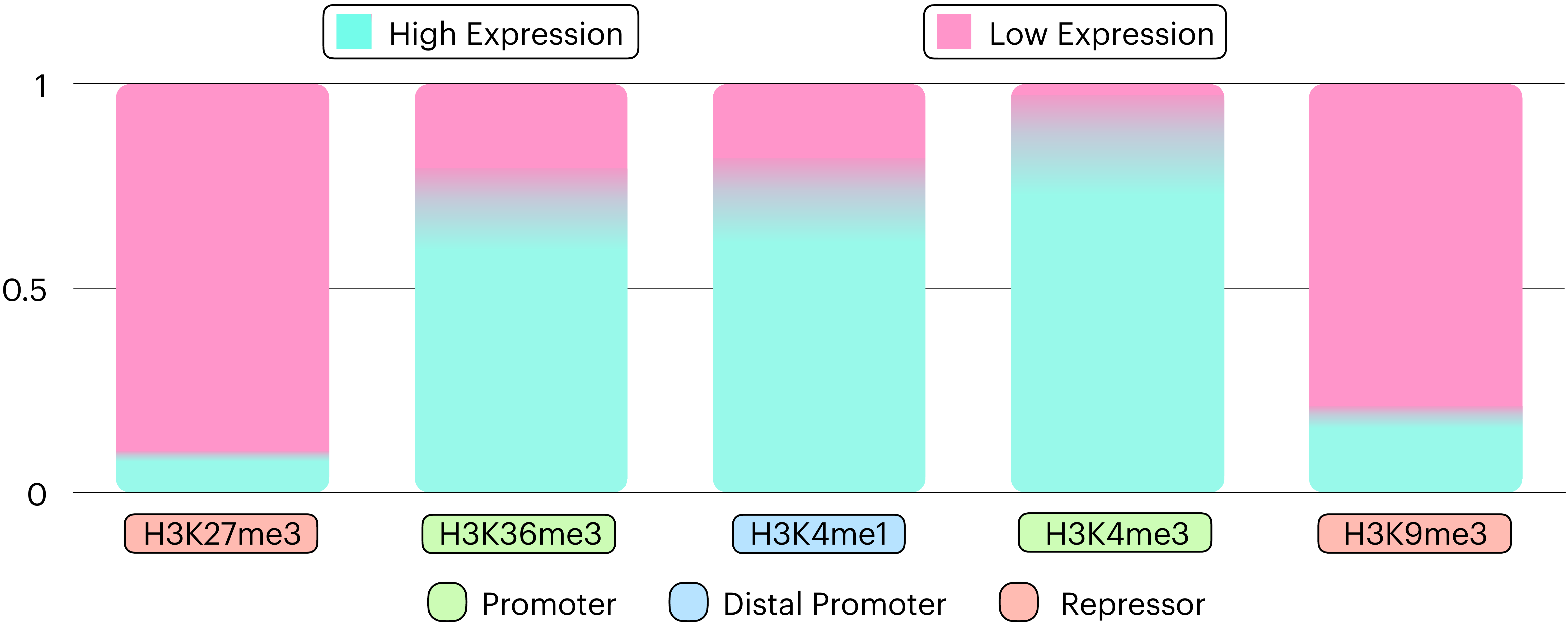}
\vspace{-8pt}
\caption{\textbf{GAN Visualization Technique Results.} Normalized activation frequency for top 100 inputs in terms of positive and negative class probability from dataset of 3 million GAN-generated inputs. Both the promoter and structural histone modification marks are activated when gene expression is high, and we observe an opposite trend with repressor marks when gene expression is low.
 }
\label{fig:gan-results}
\end{figure}

\begin{figure*}[t!]
  \centering
    \includegraphics[width=.99\textwidth]{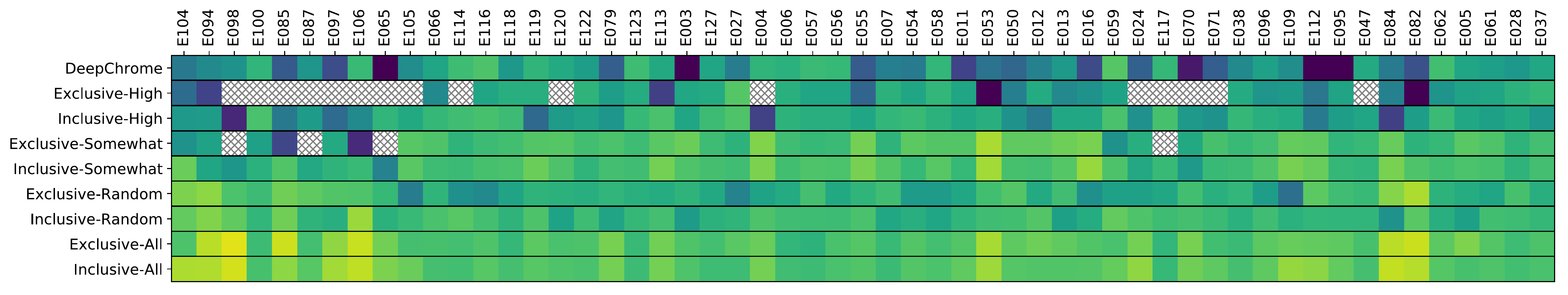}
\vspace{-6pt}
\caption{\textbf{All Cross-Cell Results.} The vertical axis represents experiment type, and the horizontal axis represents cell type. Each entry in the heatmap is the normalized AUROC of that experiment with that test cell. Brighter cells correspond to higher AUROC scores. Blank entries represent instances where no cells were well enough correlated to the test cell to run the experiment. Since the rows at the bottom are brighter, we see that training on multiple cells helps.
 }
\label{fig:cross_cell_all}
\end{figure*}

\subsection{Visualization Results}
\label{subsec:results-viz}
To generate inputs that both maximally activate a DeepChrome-style classifier and are realistic HM signals, we train a classifier and a GAN using the DeepChrome dataset, and sample the generator from the GAN to find inputs that maximize the classifier's confidence for a chosen class label. This visualization technique allows us to both confirm the network learned something reasonable and gain new insights into what the model learned.

First, we need to ensure the original data and GAN-generated data are indistinguishable to the classifier. Table~\ref{tab:gan-sanity} shows the data from the GAN is classified with the same mean confidence as the data from the real dataset. This acts as a sanity check, as unlike images, we cannot visually confirm if the GAN is producing reasonable inputs. Next, we generate 3 million inputs using the GAN, and find the top 100 inputs in terms of positive class probability and negative class probability. We then take the mean of the inputs along the temporal axis, and normalize the frequency. As seen in Fig.~\ref{fig:gan-results}, both the promoter and structural HM marks are activated when gene expression is high, and we observe an opposite trend with repressor marks when gene expression is low. From here, through generating realistic samples that extract what our models are learning, perhaps it is more feasible to learn novel insights into combinatorial relationships among HMs for gene regulation.

\subsection{Cross-cell Prediction}
\label{subsec:results-cross}

\begin{figure}[t]
  \centering
      \begin{subfigure}[b]{0.42\textwidth}
         \centering
         \includegraphics[width=1\textwidth]{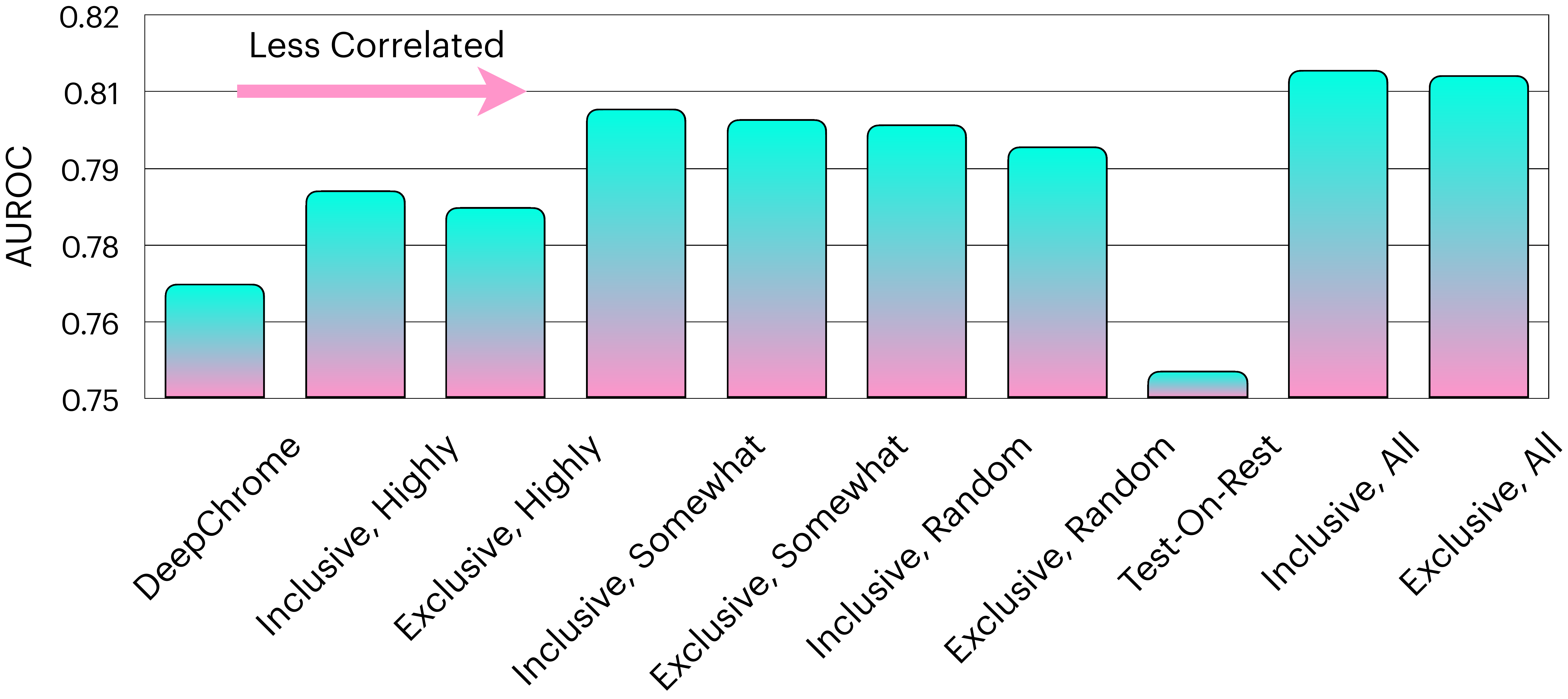}
         \caption{Sorted by correlation between training and testing data.\vspace{4pt}}% in respect to average gene expression.}
         \label{fig:cross_cell_bars_correl}
     \end{subfigure}
      \begin{subfigure}[b]{0.42\textwidth}
         \centering
         \includegraphics[width=1\textwidth]{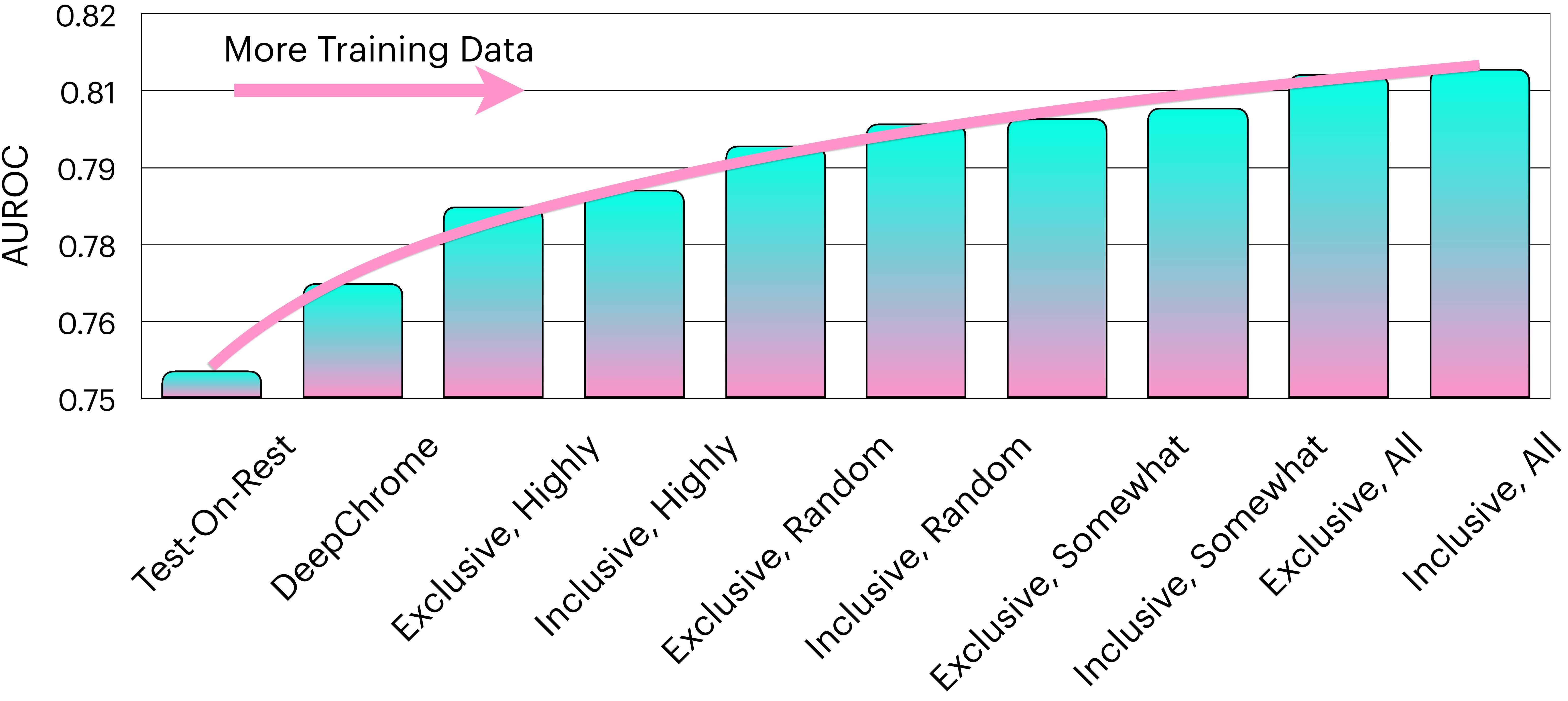}
         \caption{Sorted by training dataset size.}
         \label{fig:cross_cell_bars_dset}
     \end{subfigure}
\vspace{-8pt}
\caption{\textbf{Cross Cell Prediction Results.} Each bar measures the average AUROC of the particular experiment type over all test cells. In (a), we see no obvious relationship between the performance of the model and how correlated the cells in the training and testing sets are. In (b), we see a trend between the amount of data we train on and the AUROC.
 }
\label{fig:cross_cell_bars}
\end{figure}

\begin{figure}[t]
  \centering
    \includegraphics[width=.48\textwidth]{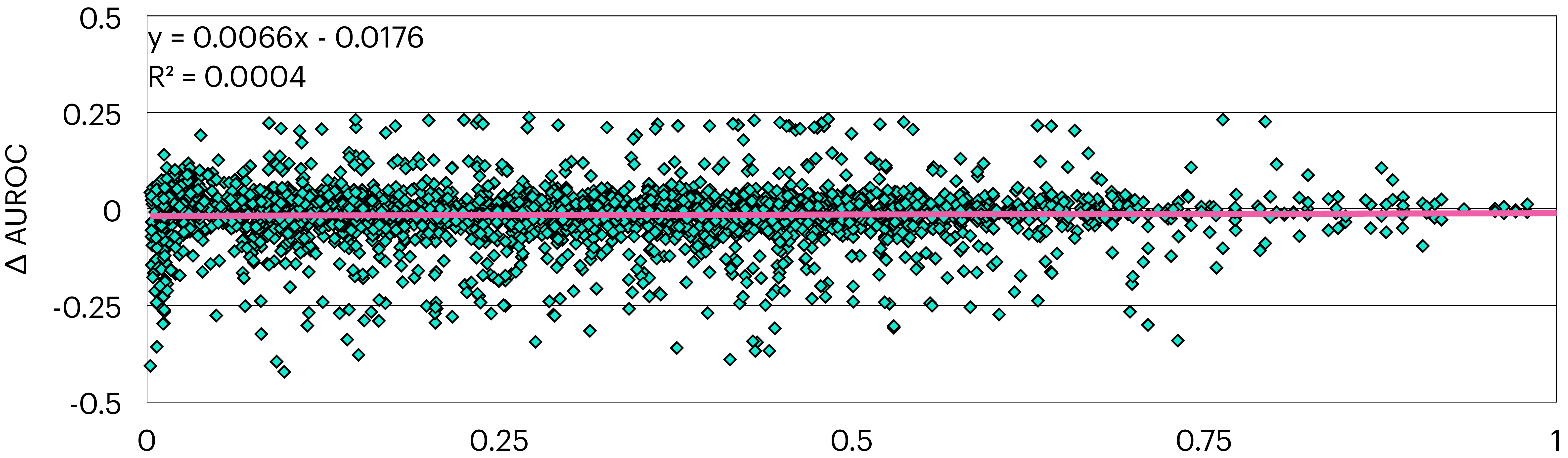}
\vspace{-16pt}
\caption{\textbf{Test-on-Rest Results.} For every possible pair of train and test cells, we plot the correlation between the train and test cell on the X-axis, and the change in AUROC (between training on the new cell versus training on the test cell) on the Y-axis.
 }
\label{fig:cross_cell_test_on_rest}
\end{figure}

Fig.~\ref{fig:cross_cell_all} contains all of our experimental results. Qualitatively, the rows at the bottom are brighter than the first row (baseline DeepChrome). Fig.~\ref{fig:cross_cell_bars_correl} contains results from Fig.~\ref{fig:cross_cell_all}, averaged across cell type, with bars sorted by the correlation between the training and test data. We expect that adding additional training data from cells that are less correlated to the test cell would actually hurt performance, and the experiments such as the Inclusive-Highly or the Inclusive-Somewhat would do the best. However, we see this is not the case. Fig.~\ref{fig:cross_cell_bars_dset} contains the same results, with bars organized by the size of the training dataset. We see a more obvious trend here: adding more data seems to increase performance, even if the data we add is not from a cell with highly correlated gene expression values to those of the test cell. Fig.~\ref{fig:cross_cell_bars_dset} also shows that training on all the cells at once results in the best performance.

Since these results seem to imply that the correlation between the training and testing cells does not impact performance, we employ our Test-on-Rest experiments to ablate this. Fig.~\ref{fig:cross_cell_test_on_rest} contains results from this set of experiments. If any substantial positive relationship existed between correlation and cell-transferability, we would expect a trendline with a strong, upwards slope. However, the trend line has a small slope and an insignificant $r$ value, illustrating that there is essentially no relationship between the change in performance when using cross-cell prediction and the correlation between the train and test cell. This result suggests that the effectiveness of cross-cell prediction is not dependent on the gene expression correlations across cells.

\begin{table}[t]
\caption{
\small
\textbf{Architectural Modifications}. Our architectural modifications made minimal differences in terms of the AUROC. Although the original paper reports an AUROC of 0.8 for the standard DeepChrome setup, we report results from our PyTorch re-implementation.\\}
\centering
\small
\begin{tabular}{| c || c | c |}
\hline 
Model Architecture & Num Parameters & Avg. AUROC
\\ \hline\hline 
Original Architecture & 645,000 & 0.77
\\ \hline
Averagepool & 645,000 & 0.78
\\ \hline
Strided Convolutions & 360,000 & 0.76
\\ \hline
Simple Linear Network & 12 & 0.77
\\ \hline
\end{tabular}
\vspace{2pt}
\label{tab:arch-mod}
\end{table}

\subsection{Linear Baseline}
\begin{figure}[t]
  \centering
    \includegraphics[width=.48\textwidth]{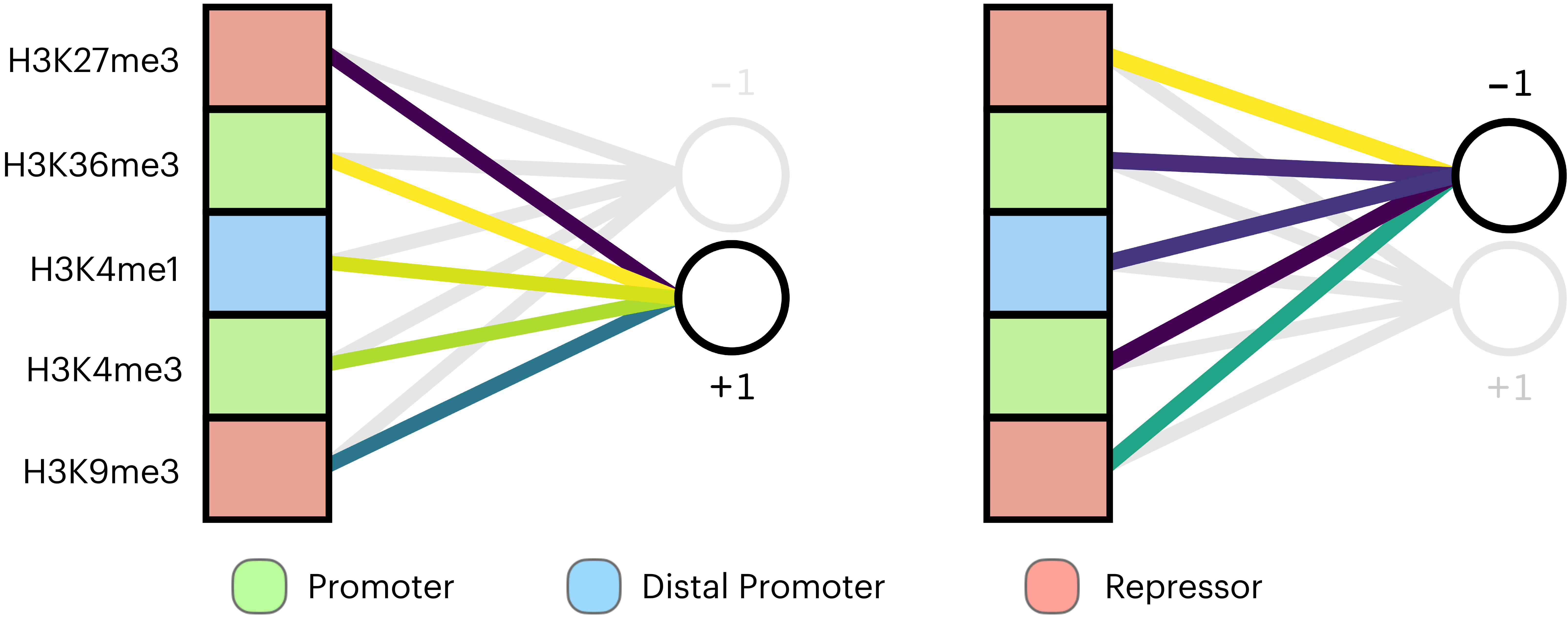}
\vspace{-16pt}
\caption{\textbf{Linear Model Weight Analysis.} Connections colored by learned weights, with brighter colors representing larger positive weights. The figure on the left shows weights corresponding to active promoters leading to high gene expression. Similarly, the figure on the right shows weights corresponding to active repressors leading to low gene expression.
 }
\label{fig:linear-weights}
\end{figure}

As we see in Table~\ref{tab:arch-mod}, changing architectures seems to make little difference in performance. This suggests the models are possibly not leveraging the benefits of each method. More concretely, in the visualization experiments, we observe that the models learn that promoters are active when gene expression is high and repressors are active when gene expression is low. Interestingly, the linear model reported an average AUROC that matched the DeepChrome AUROC of 0.77. These results suggest signals learned by DeepChrome are only as informative as the average HM counts. We analyzed what the linear network was learning in Fig.~\ref{fig:linear-weights}, which reinforced the belief that an increase in promoter counts leads to high gene expression and an increase in repressor counts leads to low gene expression.

\subsection{Investigating Independence of Cell Type on HM signals \& Gene Expression}

\begin{figure}[t]
  \centering
    \includegraphics[width=.48\textwidth]{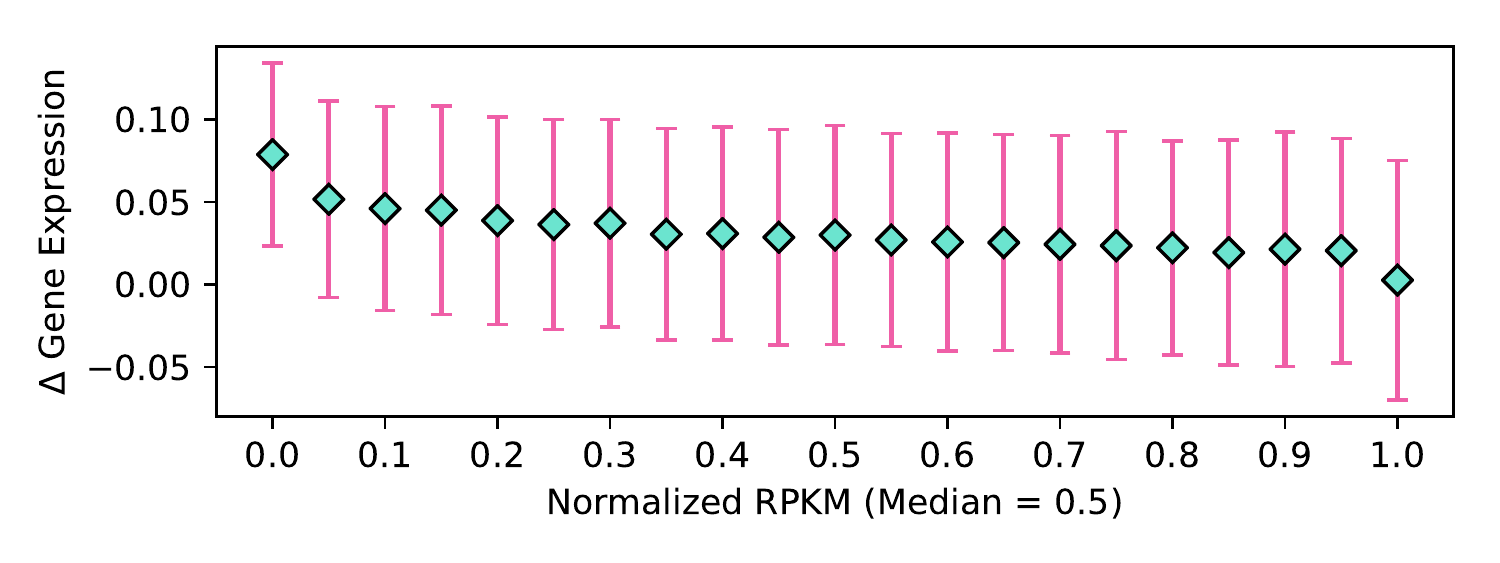}
\vspace{-24pt}
\caption{\textbf{Difference in Prediction Score Between DeepChrome and Inclusive-All Models, Binned by RPKM.} The DeepChrome classifier predicts a higher probability of gene expression for low RNA-seq RPKM counts, perhaps explaining why Inclusive-All attains a higher AUROC than DeepChrome. There does not appear to be a substantial shift in the predictions in terms of mean or variance for genes with RNA-seq RPKM counts near the binarization cutoff (seen at $0.5$ above).
 }
\label{fig:independence}
\end{figure}

Results in \ref{subsec:results-viz} and \ref{subsec:results-cross} as well as Fig.~\ref{fig:cross_cell_bars} and Fig.~\ref{fig:cross_cell_test_on_rest} all suggest the relationship between HM signals and gene expression is independent of cell type. An important thing to note is that each gene’s expression has been binarized based on the median expression of that gene, in that cell type. Thus, to substantiate this claim, we train a model on %with 
all the cell types and train a model on just one cell type, then compare performance based on the normalized RNA-seq RPKM counts. In theory, since gene expression and RNA-seq RPKM counts are correlated, genes closer to the median gene expression could have noisy labels. Thus, increasing the amount of training data could help classification near this boundary, giving the models with more training data an advantage over the DeepChrome models. However, results in Fig.~\ref{fig:independence} suggest that the Inclusive-All model does not perform substantially better than the DeepChrome model near the median boundary, but rather performs better when RNA-seq RPKM counts are low. That is, the Inclusive-All model predicts low gene expression more confidently.

\section{Discussion}

We present evidence explaining why DeepChrome's visualization technique may not result in realistic and meaningful representations of what the deep network learns. Therefore, we introduce a novel visualization technique for understanding what a DeepChrome classifier learns. To do so, we utilize a GAN to generate inputs that maximally activates our classification models. In addition, through our cross-cell prediction experiments, we demonstrate that adding more training data, even from cells that were not strongly correlated to the test cell, helped increase performance. This leads us to believe that the only knowledge that DeepChrome is able to extract about HM signals is high-level, global information. Our linear baseline, which uses 4 orders of magnitude fewer parameters, achieves similar AUROCs as DeepChrome does, suggesting that the deep learning framework does not gain more insight beyond average HM counts. Given our results suggest that the mapping between HM signals and gene expression are independent of cell type, there is likely some confounding factors affecting our results, since literature suggests otherwise.

For future work, we would like to explore if models trained on genes expressed across many cell types perform better on other widely expressed genes compared to genes that are expressed in a few cell types. This would provide insight towards understanding if HMs are cell-type or gene-type specific.

In summary, our work provides orthogonal developments to DeepChrome and its successors. With novel intuition and results gained by improvement works such as ours, we come closer to understanding and alleviating genetic diseases. \balance
 
\newpage

\bibliography{refs}
\bibliographystyle{icml2019}

\end{document}